\documentclass[letterpaper, 10 pt, conference]{ieeeconf}  
\usepackage{graphicx}
\graphicspath{{figures/}} 
\usepackage{caption}
\usepackage{subcaption}
\usepackage{url}
\usepackage{multirow}
\usepackage{tikz}

\IEEEoverridecommandlockouts                              

\title{\LARGE \bf
Single and bi-layered 2-D acoustic soft tactile skin (AST2)}

\author{Vishnu Rajendran S$^{1}$, Simon Parsons$^{1}$ and Amir Ghalamzan E.$^{2}$
\thanks{$^{1}$University of Lincoln, Lincoln, UK
}\thanks{$^{2}$University of Surrey, Computer Science, Guildford, UK
}}

\begin{document}

\maketitle
\thispagestyle{empty}
\pagestyle{empty}

\begin{abstract}
This paper aims to present an innovative and cost-effective design for Acoustic Soft Tactile (AST) Skin, with the primary goal of significantly enhancing the accuracy of 2-D tactile feature estimation.
The existing challenge lies in achieving precise tactile feature estimation, especially concerning contact geometry characteristics, using cost-effective solutions.
We hypothesise that by harnessing acoustic energy through dedicated acoustic channels in 2 layers beneath the sensing surface and analysing amplitude modulation, we can effectively decode interactions on the sensory surface, thereby improving tactile feature estimation.
Our approach involves the distinct separation of hardware components responsible for emitting and receiving acoustic signals, resulting in a modular and highly customisable skin design. Practical tests demonstrate the effectiveness of this novel design, achieving remarkable precision in estimating contact normal forces (MAE $<$ 0.8 N), 2D contact localisation (MAE $<$ 0.7 mm), and contact surface diameter (MAE $<$ 0.3 mm).
In conclusion, the AST skin, with its innovative design and modular architecture, successfully addresses the challenge of tactile feature estimation. The presented results showcase its ability to precisely estimate various tactile features, making it a practical and cost-effective solution for robotic applications.
\end{abstract}

\section{Introduction}
The widespread use of tactile sensing in manipulating (soft) objects highlights its significance. Soft tactile sensors, especially when paired with tactile prediction~\cite{Mandil2022RSS} and predictive slip control~\cite{nazari2023proactive, nazari2023deep}, minimise (soft) object damage during manipulative movements by enabling gentle interactions and providing essential feedback. Capturing crucial attributes like contact forces, shape, deformation, location, and material properties is essential for advanced control systems, shaping the execution of manipulation tasks~\cite{shuvo2022electronic}. For a detailed exploration, refer to the recent comprehensive review on tactile sensing technology~\cite{mandil2023tactile}.


In recent decades, significant strides have been taken in the advancement of soft tactile sensors, extensively documented through the use of various materials and transduction techniques~\cite{mandil2023tactile, Roberts2021SoftTS}. However, a noticeable gap persists in creating soft sensors that are easy to prototype, cost-effective, and customisable. These sensors become particularly crucial when adapting to diverse shapes of a robot's end-effectors or components. The primary challenge in customising soft tactile sensors resides in their transduction principle, translating contact properties into measurable material or medium properties, such as electrical voltage or current measurements, which serve as the foundation for inferring tactile features.

For instance, soft sensors based on electronic or electromagnetic transduction methods necessitate intricate circuitry embedded beneath the sensing surfaces~\cite{zimmer2019predicting, li2016wide, song2019pneumatic, rehan2022soft, diguet2022tactile}. Modifying the sensor's form factor to accommodate different shapes requires the fabrication of tailored circuitry to align with these alterations. Similarly, sensors employing camera-based transduction methods face the challenge of accommodating the camera unit and its associated components behind the sensing surface~\cite{ward2018tactip, yuan2017gelsight, lambeta2020digit}. This limitation constrains how much the sensor's form factor can be minimised.

\begin{figure}[tb!]
      \centering
        \includegraphics[width=0.48\textwidth]{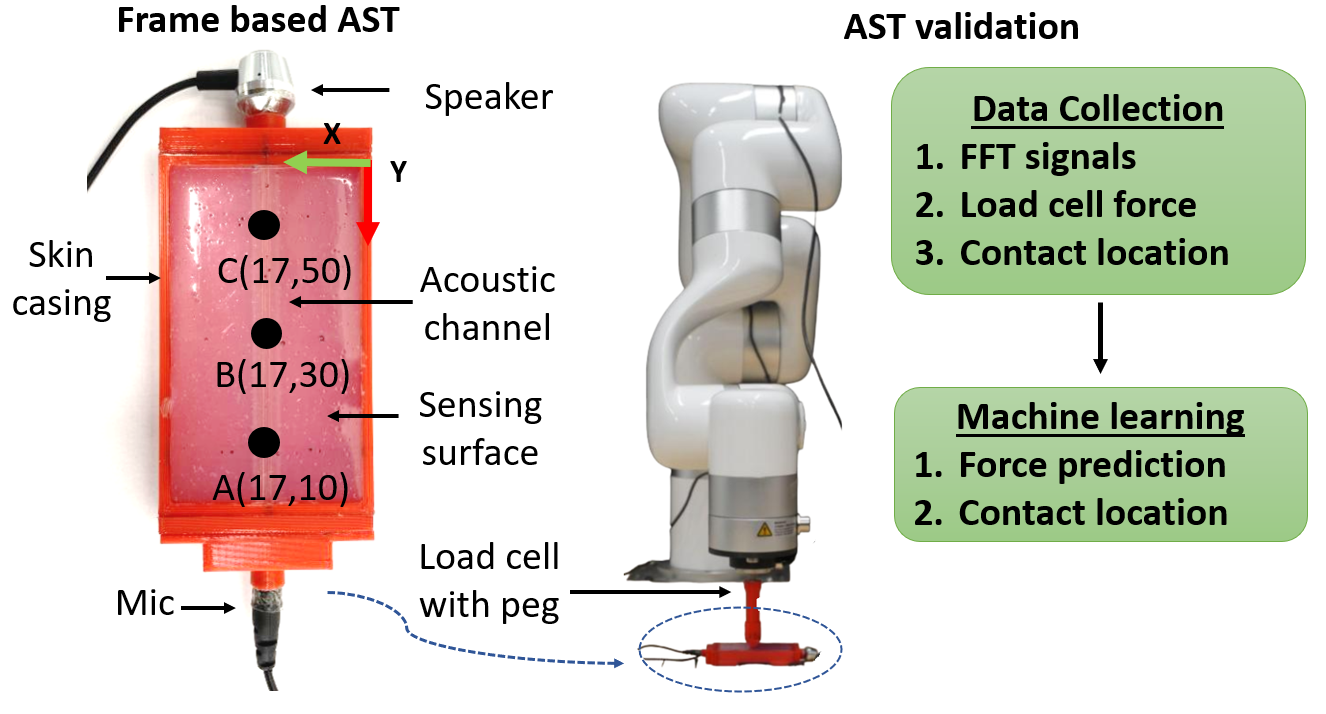}
      \caption{Initial validation of the frame-based AST sensor with cylindrical channel with three sensing points at 20 mm spatial resolution (left), robot-based calibration setup (right)~\cite{vishnu2023acoustic}.}
      \label{AST}
\end{figure}   

\begin{figure*}[tb!]
    \centering
  \subfloat[\label{Overview1}]{%
       \includegraphics[width=0.36\linewidth]{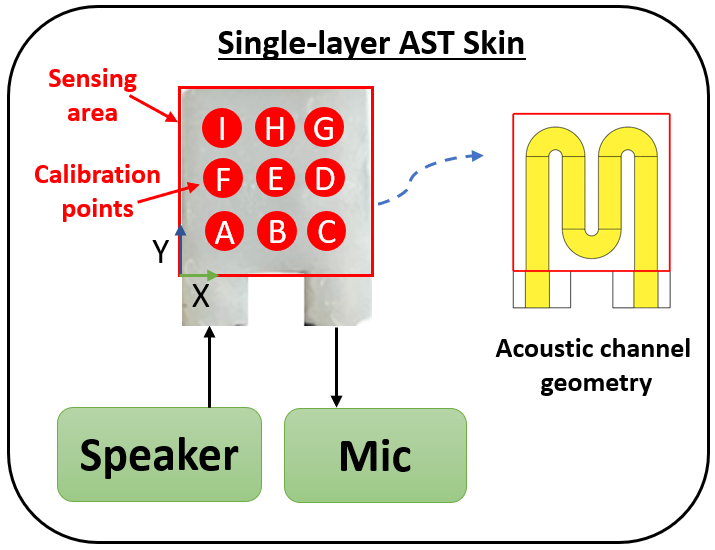}}
  \subfloat[\label{Overview2}]{%
        \includegraphics[width=0.63\linewidth]{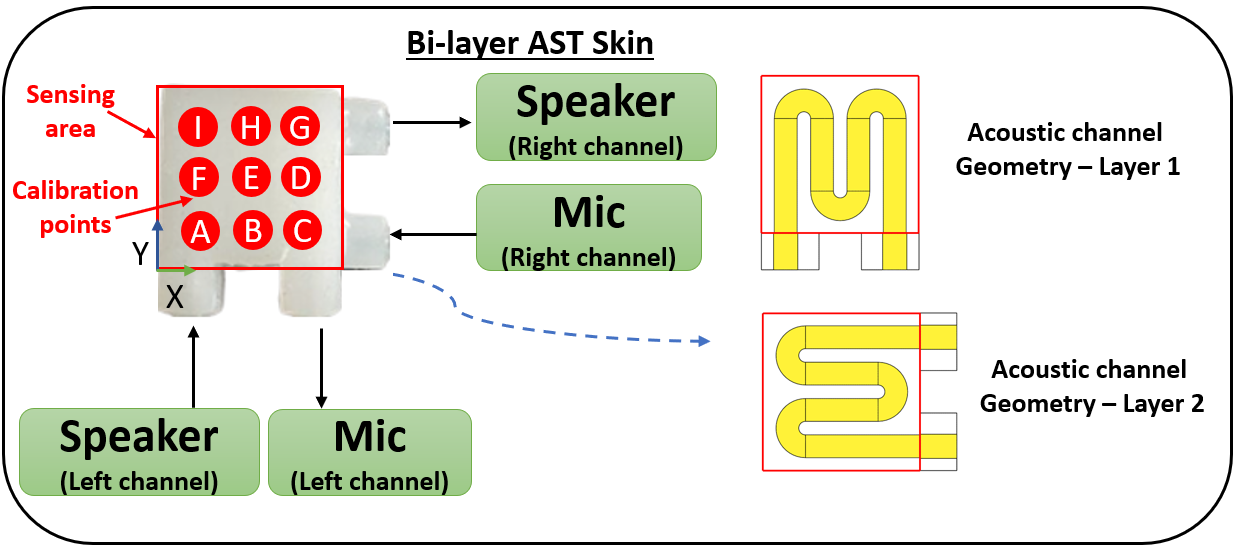}}
        
        \subfloat[\label{Overview3}]{%
        \includegraphics[width=0.7\linewidth]{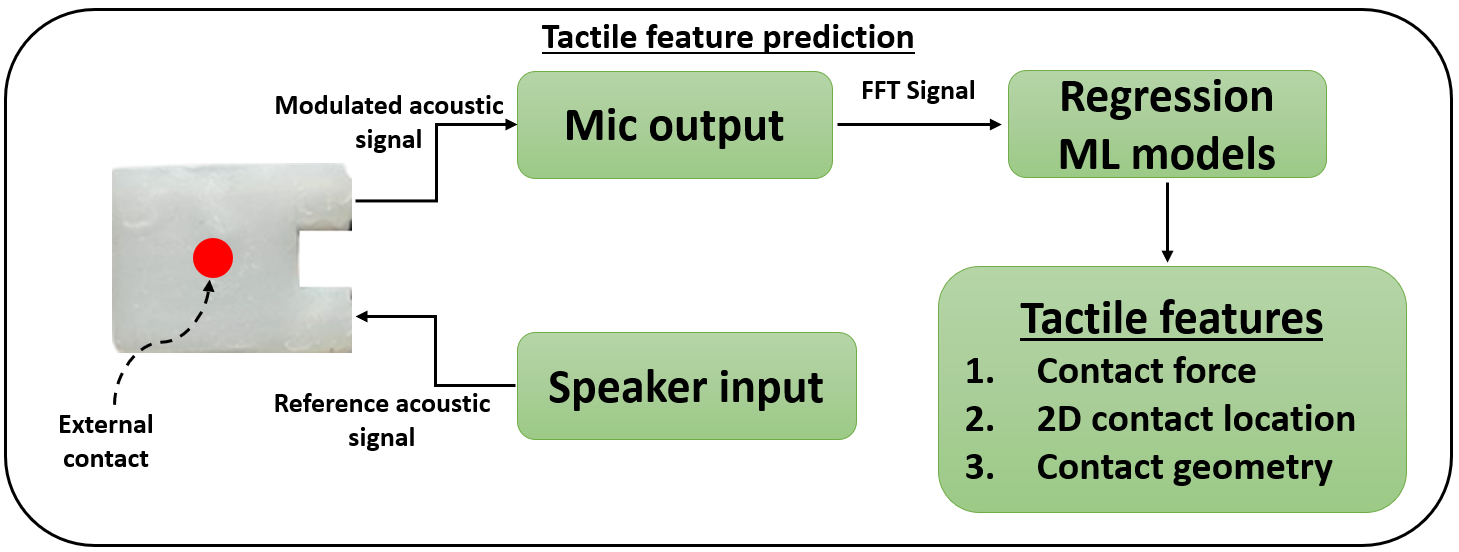}}
   \caption{AST Skin Configuration: (a) Single-layered skin (b) Bi-layered skin (c) Tactile Feature Prediction Model; Both skin designs are calibrated with calibration points at a 3 mm spatial resolution, identified by the following coordinates: \textsl{A=\{10,10\}, B=\{13,10\}, C=\{16,10\}, D=\{16,13\}, E=\{13,13\}, F=\{10,13\}, G=\{16,16\}, H=\{13,16\}, I=\{10,16\}}}
   \label{Overview}
\end{figure*} 
A pivotal approach to streamline sensor customisation involves the separation of transduction units from the sensing surface, maintaining them as distinct entities. To accomplish this, leveraging a propagating medium, such as acoustic waves or fluids, to convey information about the deformation of the sensing surface to the transduction unit is highly effective. Among these options, the acoustic method presents several advantages, primarily involving minimal hardware components for the transduction unit. This concept was demonstrated by sensorising a soft pneumatic actuator, which derived tactile features using a speaker and microphone embedded inside the pneumatic actuator~\cite{zoller2018acoustic,wall2022virtual,zoller2020active}. Moreover, there is another reason to prefer acoustics over fluid-based methods. This is due to the fact that fluid-based systems are known for their delayed response to tactile interactions~\cite{fujiwara2023agar}. Moreover, the acoustics method has been used by certain sensors to derive tactile features like forces~\cite{chuang2018ultrasonic}, contact locations~\cite{351130}, resulting deformations~\cite{shinoda1996tactile}, and contact shapes~\cite{chuang2018ultrasonic,shinoda1997acoustic}. However, these mentioned sensors have embedded electronic components below the sensing surface to emit acoustic waves and sense the returning signals, and that limits the possibility of customisation without altering the electronics. As per our surveys, we couldn't find any such tactile skin technology without embedded electronic circuits that provide room for easy customisation. A low-cost Acoustic Soft Tactile Skin (AST Skin) has been developed to address some of these shortcomings. AST can be attached to flat~\cite{rajendran2023towards} and non-flat surafaces~\cite{vishnu2023acoustic}. 

AST skin utilises acoustic waves travelling through the acoustic channels positioned beneath the sensing surface to transfer the effects of sensor deformation to the transduction unit. AST does not need to embed electronic hardware near the soft sensing surface. Instead, it allows the possibility of positioning the acoustic transduction components at convenient locations. This feature provides the flexibility to create modular skins in various shapes and form factors as long as acoustic channels can be designed beneath the sensing surface. In \cite{vishnu2023acoustic}, the skin was solely validated for measuring contact normal forces with 1-D contact localisation. Moreover, the early stage prototypes of the AST skins were having the acoustic hardware attached to distal ends on a rigid casing (refer Fig.~\ref{AST}).  

This paper introduces the new version of the novel AST skin designed to measure contact normal forces with 2D contact localisation and contact geometry features (refer Fig.~\ref{Overview}). The key contributions of this work are: (i). Development of a modular (2-d) tactile skin design, strategically isolating the acoustic hardware components from the skin structure. (ii). Examination of the impact of incorporating multiple-layer acoustic channels on the skin's performance, thereby exploring its sensing capabilities.


\section{AST Sensing Technology}
In this section, we delve into the concept of AST sensing technology and explore the initial studies that have been undertaken to evolve into a functional soft tactile skin.
\paragraph*{Concept} AST technology represents a tactile sensing approach that leverages acoustic channels beneath the sensing surface to interpret tactile interactions occurring on the surface. The deformation of these channels effectively alters the reference sound wave that travels through them. Subsequently, this modulation is then translated into tactile readings through the utilisation of trained machine learning models. 

\paragraph*{Preliminary validation} To validate the concept of AST skin technology, the first attempt was to test two assumptions: (i) whether a sound wave propagating through a channel gets modulated when its cross-section is altered ?, and (ii) whether this modulation can be mapped to quantify the stimulus that caused the deformation (that is the force and its contact location)?

A test skin design was prototyped for testing the assumptions, as shown in figure~\ref{AST}. In this prototype, the silicon rubber (the skin) is deposited inside a 3D-printed plastic casing, leaving a cylindrical channel through the skin. A standard headphone speaker and a microphone are arranged on either side of the opening on the casing such that the microphone can directly listen to the sound propagating from the speaker.

\begin{figure}[h]
    \centering
  \subfloat[\label{1a}]{%
       \includegraphics[width=0.45\linewidth]{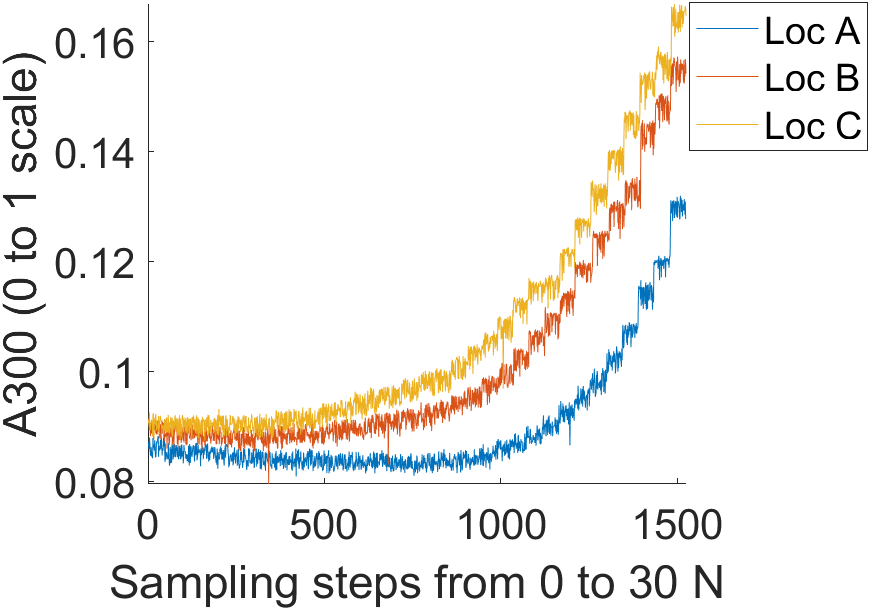}}
    \hfill
  \subfloat[\label{1b}]{%
        \includegraphics[width=0.45\linewidth]{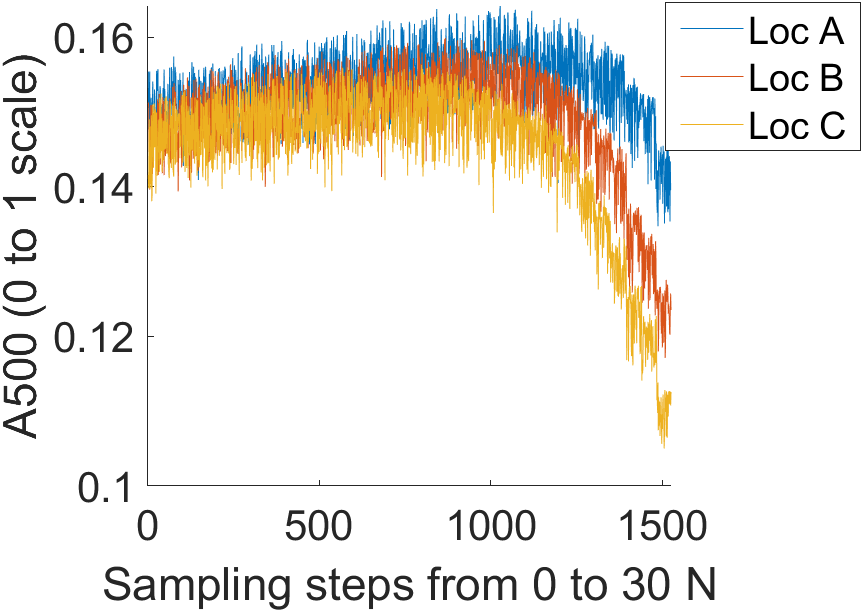}}
    \\
  \subfloat[\label{1c}]{%
        \includegraphics[width=0.45\linewidth]{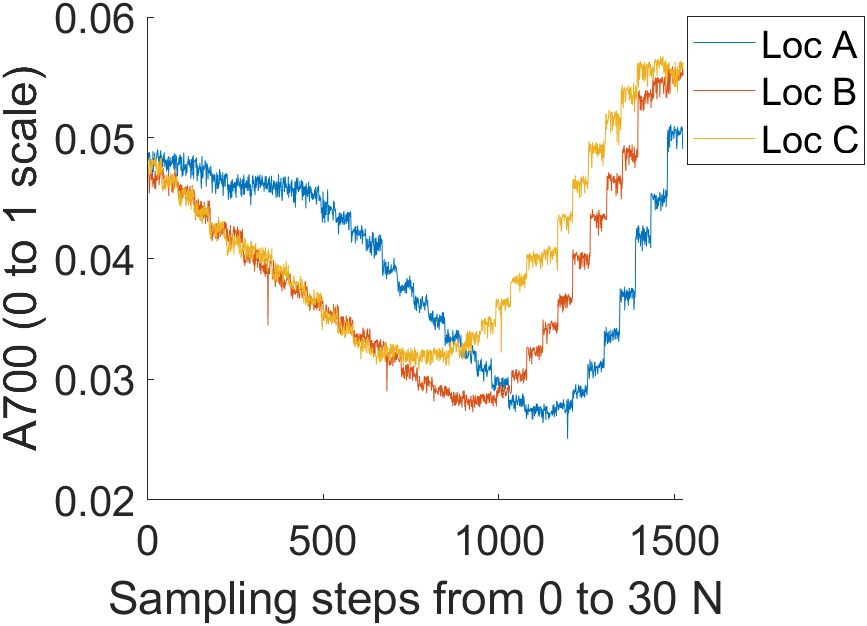}}
    \hfill
  \subfloat[\label{1d}]{%
        \includegraphics[width=0.45\linewidth]{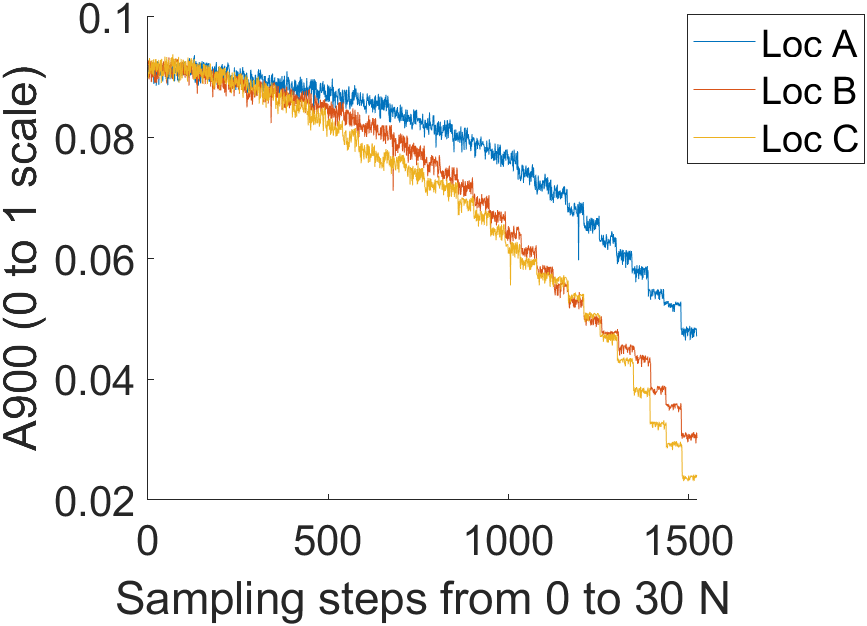}}
   \label{Amplitude}
\caption{Variation of FFT data at locations A, B, and C when force varies from 0 to 30$^{+1}$ N for (a) 300 Hz, (b) 500 Hz, (c) 700 Hz, and (d) 900 Hz.  }
\label{fftvariation}
\end{figure}

A test reference sound wave of known frequencies and amplitude is played through the headphone, and the output sound is recorded to a PC through the microphone. This test sound comprises four sine waves of frequencies 300 Hz, 500 Hz, 700 Hz, and 900 Hz, with an amplitude of 0.6 for each frequency component. A gradual normal force from 0-30  N is applied on three distinctive points (A, B, and C) on the skin's surface using a 6-degree-freedom robotic arm mounted with a load cell and peg to the wrist (refer Fig.\ref{AST}). The robot pushes the peg on the skin with equal vertical travel increments until the load cell value reaches 30 N. During each increment, the corresponding sound received by the microphone is recorded and converted to FFT data and tabulated against the applied load. This FFT data was analysed and found that the reference signal has undergone amplitude modulation as the cylindrical passage deformed due to the externally applied load (see Fig~\ref{fftvariation}). This verified our first assumption. Moreover, it is also evident that these FFT variations follow a different trend at the three locations. And this confirmed our second assumption that FFT data could be utilised as a primary predictor of the external force with their contact location, which caused the deformation of the skin. 

Later, these FFT data with their corresponding load value and contact location are used to train a regression and classifier model to test the skin's performance in predicting force and their contact location from FFT data. We have employed the Exponential Gaussian process and Bagged Ensemble Trees machine learning models for predicting contact force and location, respectively. It has been studied that the skin with the cylindrical channel can predict 95.5\% of the contact forces with $\pm$1 N tolerance and 1-D contact location with 20 mm resolution at 96.66\% accuracy. We also tested different acoustic channel geometries; the cylindrical channel was the best performer among them~\cite{vishnu2023acoustic}.

In this paper, we aim to develop and compare two frame-less skin designs for sensing contact normal forces, 2D contact location (X and Y), and contact geometry features.

\section{Single and Bi-layered AST skin}
\begin{figure}[tb!]
      \centering
        \includegraphics[width=0.39\textwidth]{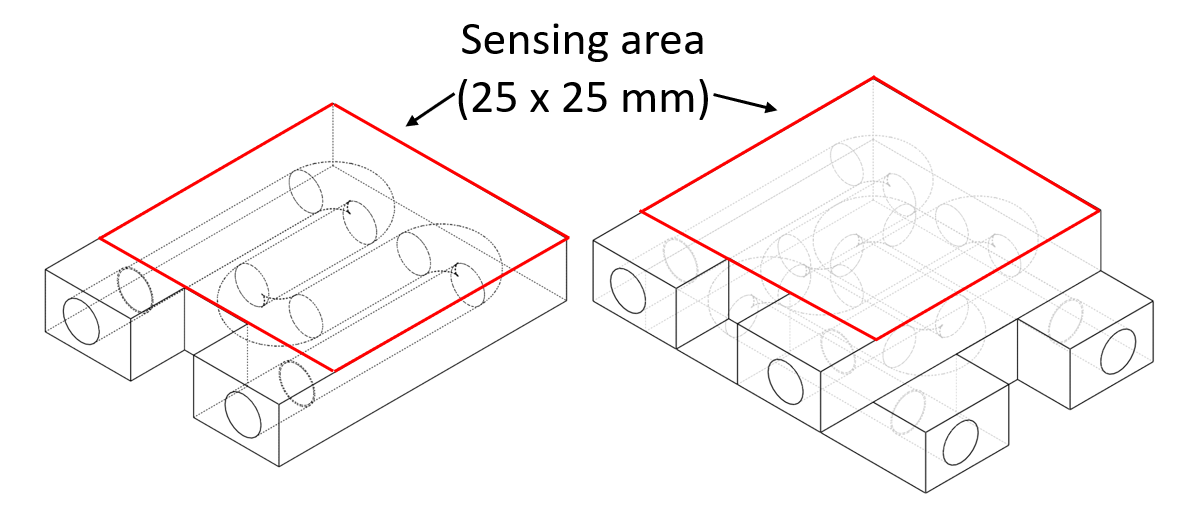}
      \caption{Skin design: (a). Single-layer (left) skin and Bi-layer skin (right)}
      \label{Design}
\end{figure}

Here we consider two test skin designs, i.e., a single-layer and a bi-layer skin with a sensing area of 25 x 25 mm (refer Fig.~\ref{Overview} and \ref{Design}). The single-layered skin has a continuous cylindrical acoustic channel of 3 mm diameter running throughout the length and breadth of the skin. In the bi-layer skin design, two single-layer skins are sandwiched orthogonally. These two design variation allows us to distinguish the sensing performance when channels are arranged in single or multiple layers below the sensing surface. Moreover, bi-layer skin allows more deformation than single-layer skin; hence, the sensing force range can be enhanced.


\begin{figure*}[h]
    \centering
  \subfloat[\label{}]{%
       \includegraphics[width=0.15\linewidth]{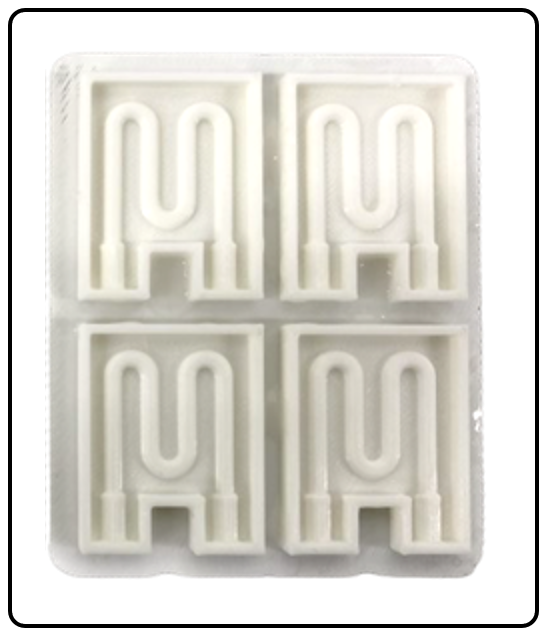}}
  \subfloat[\label{}]{%
        \includegraphics[width=0.15\linewidth]{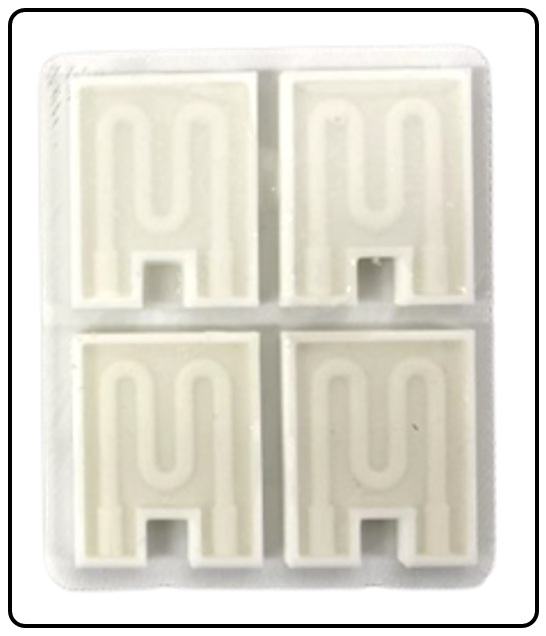}} 
  \subfloat[\label{}]{%
        \includegraphics[width=0.15\linewidth]{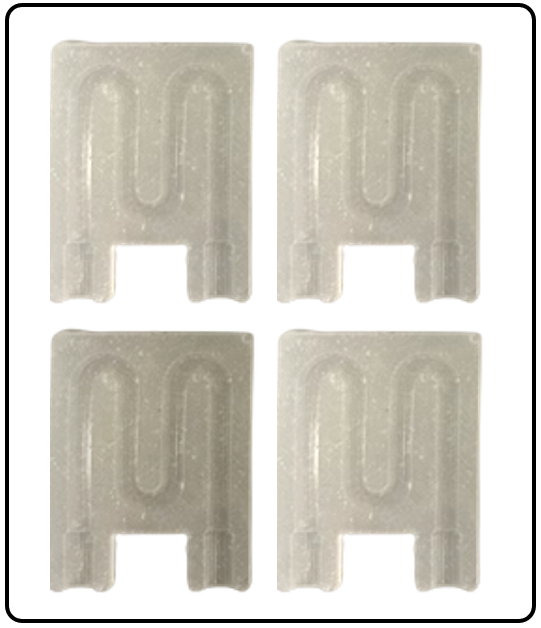}}
          \subfloat[\label{}]{%
        \includegraphics[width=0.15\linewidth]{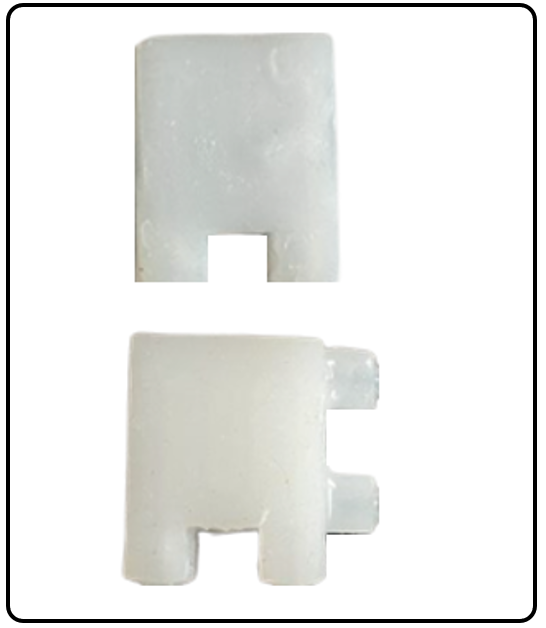}}
        \hfill
          \subfloat[\label{}]{%
        \includegraphics[width=0.28\linewidth]{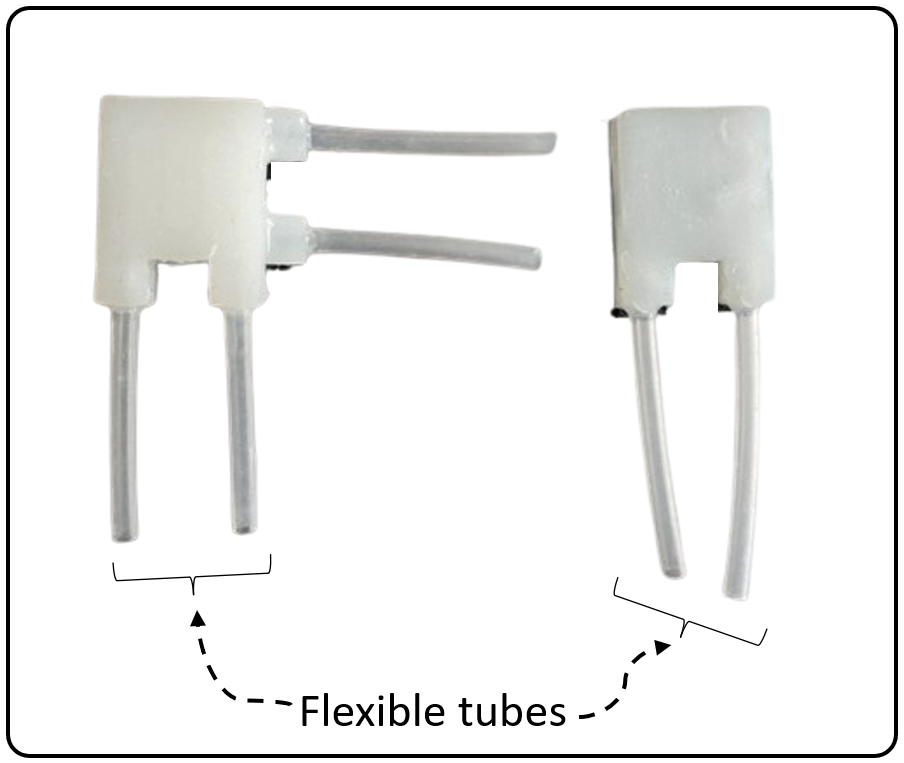}}
  \subfloat[\label{}]{%
        \includegraphics[width=0.41\linewidth]{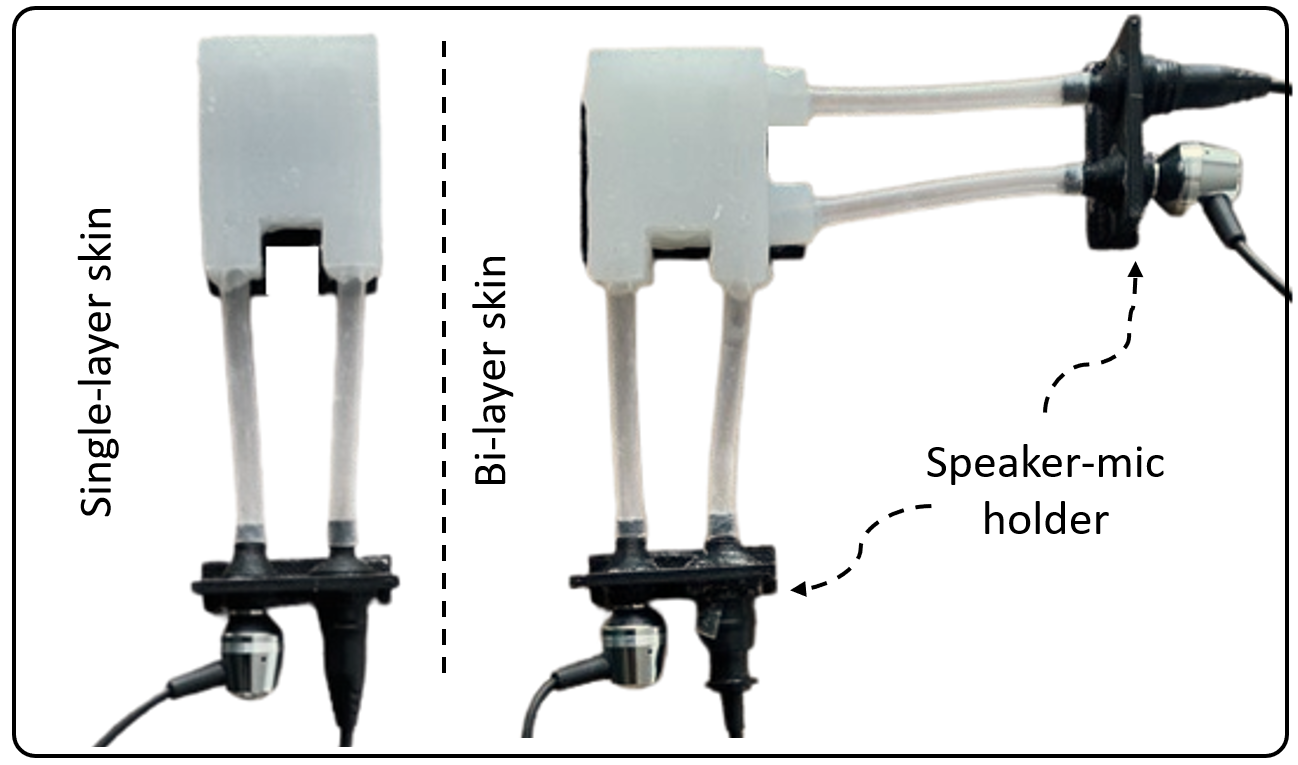}}
   \label{Prototye}
\caption{Prototyping process: (a) 3D printed moulds (b). Silicone gel mixed with the curing catalyst are poured into the mould (c). Cured skin is removed from the mould (d). Cured skin halves joined to make the single-layered skin (top), and two single-layered skin joined orthogonally to form the bi-layered skin (bottom) (e). Connecting flexible tube to extend the acoustic channels (f). Mounting the speaker and microphone on the holder and fastening to the flexible tubes}
\label{Prototyping}
\end{figure*}

\paragraph*{Prototyping process} For prototyping these skin designs, moulding techniques have been used. The required moulds are designed and 3D printed with PLA material. Silicone material with a 10 A shore hardness value (PlatSil Gel 10) is used as the skin material. The skin material is mixed with its curing catalyst at a 1:1 ratio and poured into the moulds. Two moulds are used to cast the two halves of a skin layer with a channel cavity about its mid-plane. Later these two halves are joined together to form the single-layer skin. For the multi-layer skin, two single-layer skins are sandwiched orthogonally. For joining the cured skin halves and for sandwiching the skins, a thin layer of the same silicone material with its catalyst is applied and cured. The acoustic channel ends are extended by flexible tubes to connect with the speaker and microphone. To ensure the portability of the skin during the testing purpose, the skins are mounted to a 3D printed base (Fig.~\ref{Prototyping} (e) and (f)). Standalone mountings are provided to support the speaker and microphone that are connected to the respective flexible tubes. For testing the concept, general-purpose headphone speakers and microphones are used, which will be replaced with the miniature versions. Moreover, to route the flexible tube without bending, a plastic casing can be used. This is needed when the acoustic hardware can't be placed in line with the skin ports. The prototyping procedure is presented in the figure~\ref{Prototyping}.   

\begin{figure}[h]
      \centering
        \includegraphics[width=0.3\textwidth]{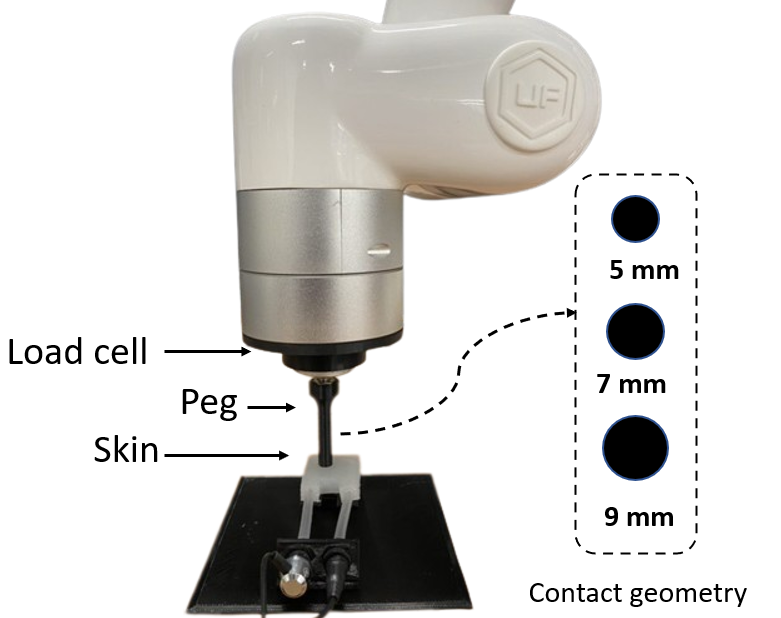}
      \caption{3D printed peg with flat-circular end used for calibrating the skin. As a test case, contact surface diameter is a feature to be predicted by the skin}
      \label{peg}
\end{figure}

\paragraph*{Skin Calibration} Contact normal forces, 2D-contact locations, and contact surface geometry are key tactile information for which the skin is calibrated. A data-driven approach has been adopted for calibrating the skin to predict the tactile information from the modulated FFT data upon skin deformation. This approach tackles the complexity of acoustic-material deformation interaction modelling and provides room for testing different skin designs, acoustic parameters, and material combinations. Moreover, this approach brings this technology a scalable sensing method that can be applied to various applications where skin needs to take different form factors.

To create the data-driven calibration model, it's essential to generate comprehensive data for each skin design. This dataset comprises modulated FFT data corresponding to the normal forces applied at various points on the sensing surface and for different contact geometries. This data has been produced using a robot manipulator-based test rig as used in the preliminary validation studies~(refer to Fig.~\ref{AST}). For the current skin designs, 3 x 3 2D calibration points are considered at a resolution of 3 mm (refer Fig.~\ref{Overview}a, b).

\paragraph*{Data collection} For data collection, the robot manipulator is driven to the selected calibration points of both the skin designs and applied normal force by incrementally pushing the peg by 0.5 mm into a single-layer skin and 1 mm for bi-layer skin. For the single-layer skin, a total peg travel of 3 mm is considered, while for bi-layered skin, it is 6 mm as the bi-layered skin is expected to provide more deformation and hence the force range. Moreover, three test pegs are used to apply the forces. These pegs have circular flat bottoms with different diameters, i.e., 5 mm, 7 mm, 9 mm (refer Fig.~\ref{peg}). For the single-layer skin, when pressed to a depth of 3 mm, the maximum force values recorded for the 5 mm, 7 mm, and 9 mm pegs are approximately 3 N, 5 N, and 7 N, respectively. Conversely, for the bi-layered skin, the maximum recorded values are 6 N, 8 N, and 10 N for the respective pegs at a depth of 6 mm. This disparity in force values for each peg at similar depths is due to the effect of the contact area on the skin. The rationale behind using these different pegs is to test the skin's ability to predict contact geometry specifications. As a test case, the peg diameter is considered to represent the contact geometry. 

During data collection, the test reference sound signal used in our preliminary study is played through the speaker (i.e., the sine wave with 300 Hz, 500 Hz, 700 Hz, and 900 Hz frequency components), and the microphone listens to the returning signal. While applying the forces with different pegs, the modulated sound signal in the form of FFT data is recorded against the corresponding load cell readings with respect to the location of the action and the peg used. This forms the comprehensive data set for developing the machine learning model for predicting the contact force with its location and contact geometry features. For each skin, a total of 16632 data has been collected.

\paragraph*{Tactile feature prediction model} A Regression machine learning technique is employed to predict the tactile features for both skins. This technique trains four individual regression models to predict each feature separately from the respective FFT data. Various regression models (Linear, Trees, Support Vector Machine, Gaussian Process, Neural Network) are compared using Matlab Regression Learner to pick the best-suited models for each feature prediction for both skin designs. For this comparison, the calibration dataset is used at a 90:10 partition (with 25\% holdout for validation) for training and testing the models. The Exponential Gaussian Process Regression has been selected as the best based on the validation error (Root Mean Square Error-RMSE). The respective validation errors of each tactile feature prediction model are presented in table~\ref{MLmodel}. 

\section{Results and discussion}
\paragraph*{Calibration model performance} The tactile feature prediction results for both skin designs are presented in table~\ref{Error}. The results show that both skins could predict the contact normal forces with a mean error of less than 0.10 N, provided single-layer skin performed slightly better than the bi-layer sensor. For other tactile parameters, like peg diameter and contact locations, the bi-layered sensor outperforms the single-layered sensor.  The mean absolute error (MAE) for predicting peg diameter and contact X, Y location are 0.10 mm, 0.21 mm, and 0.20 mm for bi-layer skin, while it is 0.30 mm, 0.51 mm, and 0.61 mm for single-layer skin. 
\begin{table}[tb!]
\centering
\caption{Validation error for each Exponential Gaussian Process Regression Model}
\label{MLmodel}
\begin{tabular}{|l|cc|}
\hline
\multirow{2}{*}{\textbf{Model  and response}} & \multicolumn{1}{c|}{\textbf{Single layer   skin}} & \textbf{Bi-layer skin} \\ \cline{2-3} 
                                              & \multicolumn{2}{c|}{\textbf{Validation Error}}                             \\ \hline
Model 1: Force (N)                                & \multicolumn{1}{c|}{0.12}                       & 0.23               \\ \hline
Model 2: Peg diameter (mm)                         & \multicolumn{1}{c|}{0.66}                      & 0.31                   \\ \hline
Model 3: Location X (mm)                           & \multicolumn{1}{c|}{1.04}                      & 0.56                   \\ \hline
Model 4: Location Y (mm)                           & \multicolumn{1}{c|}{1.11}                      & 0.52                   \\ \hline
\end{tabular}
\end{table}

\begin{figure}[tb!]
      \centering
        \includegraphics[width=0.14\textwidth]{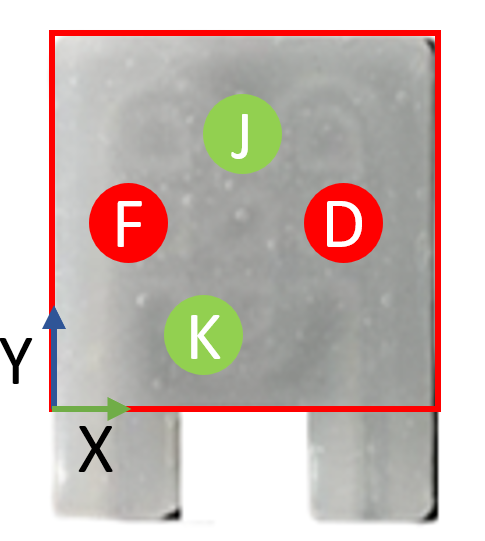}
      \caption{Selected points for skin testing: Calibrated points (Red) \textsl{D=\{16,13\}}, \textsl{F=\{10,13\}};  Non-calibrated points (Green): \textsl{K=\{10,12\}}, \textsl{J=\{13,14\}}}
      \label{test}
\end{figure}

\begin{table}[tb!]
\centering
\caption{Mean Absolute Error (MAE) in predicting the tactile features during calibration}
\label{Error}
\begin{tabular}{|l|cccc|}
\hline
\multirow{3}{*}{\textbf{Response}} & \multicolumn{4}{c|}{\textbf{Prediction   Error}}                                                        \\ \cline{2-5} 
                                   & \multicolumn{2}{c|}{\textbf{Single   layer skin}}       & \multicolumn{2}{c|}{\textbf{Bi-layer   skin}} \\ \cline{2-5} 
                                   & \multicolumn{1}{l|}{Mean}  & \multicolumn{1}{c|}{STDEV} & \multicolumn{1}{c|}{Mean}        & STDEV      \\ \hline
Force (N)                          & \multicolumn{1}{c|}{0.06} & \multicolumn{1}{c|}{0.12} & \multicolumn{1}{c|}{0.07}       & 0.18      \\ \hline
Peg diameter   (mm)                & \multicolumn{1}{c|}{0.30} & \multicolumn{1}{c|}{0.50} & \multicolumn{1}{c|}{0.10}        & 0.26        \\ \hline
Location X   (mm)                  & \multicolumn{1}{c|}{0.51} & \multicolumn{1}{c|}{0.77} & \multicolumn{1}{c|}{0.21}       & 0.51       \\ \hline
Location Y   (mm)                  & \multicolumn{1}{c|}{0.61} & \multicolumn{1}{c|}{0.82} & \multicolumn{1}{c|}{0.20}        & 0.41      \\ \hline
\end{tabular}
\end{table}
\paragraph*{Real-time sensing performance} An experimental study is conducted to assess the performance of skin designs in real-time. This study used the same calibration setup to apply a known force to two calibrated and two non-calibrated points on the skin surface. These test locations are illustrated in the figure~\ref{test}. Using three pegs, a sample test force of 3 N is applied to both skins at these selected locations for 15 trials. 3N is the maximum force that can be applied to the present single-layer skin with a 5 mm peg, and hence it is selected as a common test force for all skins for result uniformity. Subsequently, the skin's output is read in real-time after fine-tuning and is compared with their respective ground truth values. The MAE obtained during the trials is presented in the table~\ref{res}.

The results show that both types of skin designs are capable of accurately measuring tactile features at both calibrated and non-calibrated points, with errors that are within an acceptable range. However, when comparing the performance of the two types of skin, it becomes evident that the bi-layered skin outperforms the single-layered skin in measuring peg diameter and contact location in the X and Y axes, with maximum MAE of 0.28 mm, 0.49 mm, and 0.69 mm, respectively. Notably, these maximum errors are observed at non-calibrated points for the bi-layered skin. Conversely, the single-layered skin exhibits a higher maximum error in peg diameter and location in the X and Y axes, with values of approximately 1.08 mm, 1.57 mm, and 1.53 mm, respectively. In the case of contact force measurement, the bi-layered skin performs slightly lower than the single-layered skin with a difference in MAE of 0.07 N, which is negligible. But while looking at the overall performance, the bi-layered skin can be taken as the best performer.

\begin{table*}[]
\centering
\caption{Mean Absolute Error (MAE) while predicting the tactile features real-time}
\label{res}
\begin{tabular}{|c|c|cccc|cccc|}
\hline
\multirow{2}{*}{\textbf{\begin{tabular}[c]{@{}c@{}}Test\\ points\end{tabular}}} & \multirow{2}{*}{\textbf{Peg}} & \multicolumn{4}{c|}{\textbf{Single-layer skin}}                                                                                           & \multicolumn{4}{c|}{\textbf{Bi-layer skin}}                                                                                               \\ \cline{3-10} 
                                                                                &                               & \multicolumn{1}{c|}{\textbf{Force (N)}} & \multicolumn{1}{c|}{\textbf{Peg dia (mm)}} & \multicolumn{1}{c|}{\textbf{LocX (mm)}} & \textbf{LocY (mm)} & \multicolumn{1}{c|}{\textbf{Force (N)}} & \multicolumn{1}{c|}{\textbf{Peg dia (mm)}} & \multicolumn{1}{c|}{\textbf{LocX (mm)}} & \textbf{LocY (mm)} \\ \hline
\multirow{3}{*}{D \{16,13\}}                                                    & 5                             & \multicolumn{1}{c|}{0.18}               & \multicolumn{1}{c|}{1.08}                  & \multicolumn{1}{c|}{1.01}          & 0.29          & \multicolumn{1}{c|}{0.31}               & \multicolumn{1}{c|}{0.01}                  & \multicolumn{1}{c|}{0.1}           & 0.21          \\ \cline{2-10} 
                                                                                & 7                             & \multicolumn{1}{c|}{0.43}               & \multicolumn{1}{c|}{0.09}                  & \multicolumn{1}{c|}{0.86}          & 0.38          & \multicolumn{1}{c|}{0.3}                & \multicolumn{1}{c|}{0.05}                  & \multicolumn{1}{c|}{0.31}          & 0.1           \\ \cline{2-10} 
                                                                                & 9                             & \multicolumn{1}{c|}{0.21}               & \multicolumn{1}{c|}{0.2}                   & \multicolumn{1}{c|}{0.3}           & 0.76          & \multicolumn{1}{c|}{0.75}               & \multicolumn{1}{c|}{0.1}                   & \multicolumn{1}{c|}{0.1}           & 0.1           \\ \hline
\multirow{3}{*}{F \{10,13\}}                                                    & 5                             & \multicolumn{1}{c|}{0.65}               & \multicolumn{1}{c|}{0.84}                  & \multicolumn{1}{c|}{0.91}          & 0.29          & \multicolumn{1}{c|}{0.41}               & \multicolumn{1}{c|}{0.21}                  & \multicolumn{1}{c|}{0.49}          & 0.33          \\ \cline{2-10} 
                                                                                & 7                             & \multicolumn{1}{c|}{0.63}               & \multicolumn{1}{c|}{0.06}                  & \multicolumn{1}{c|}{0.74}          & 0.35          & \multicolumn{1}{c|}{0.32}               & \multicolumn{1}{c|}{0.1}                   & \multicolumn{1}{c|}{0.14}          & 0.35          \\ \cline{2-10} 
                                                                                & 9                             & \multicolumn{1}{c|}{0.68}               & \multicolumn{1}{c|}{0.06}                  & \multicolumn{1}{c|}{0.58}          & 0.37          & \multicolumn{1}{c|}{0.3}                & \multicolumn{1}{c|}{0.1}                   & \multicolumn{1}{c|}{0.2}           & 0.24          \\ \hline
\multirow{3}{*}{J \{13,14\}}                                                    & 5                             & \multicolumn{1}{c|}{0.17}               & \multicolumn{1}{c|}{0.3}                   & \multicolumn{1}{c|}{0.12}          & 0.68          & \multicolumn{1}{c|}{0.33}               & \multicolumn{1}{c|}{0.1}                   & \multicolumn{1}{c|}{0.19}          & 0.1           \\ \cline{2-10} 
                                                                                & 7                             & \multicolumn{1}{c|}{0.08}               & \multicolumn{1}{c|}{0.07}                  & \multicolumn{1}{c|}{0.13}          & 0.68          & \multicolumn{1}{c|}{0.26}               & \multicolumn{1}{c|}{0.1}                   & \multicolumn{1}{c|}{0.11}          & 0.29          \\ \cline{2-10} 
                                                                                & 9                             & \multicolumn{1}{c|}{0.1}                & \multicolumn{1}{c|}{0.58}                  & \multicolumn{1}{c|}{0.86}          & 0.6           & \multicolumn{1}{c|}{0.3}                & \multicolumn{1}{c|}{0.1}                   & \multicolumn{1}{c|}{0.11}          & 0.21          \\ \hline
\multirow{3}{*}{K \{10,12\}}                                                    & 5                             & \multicolumn{1}{c|}{0.48}               & \multicolumn{1}{c|}{0.47}                  & \multicolumn{1}{c|}{0.64}          & 1.24          & \multicolumn{1}{c|}{0.26}               & \multicolumn{1}{c|}{0.28}                  & \multicolumn{1}{c|}{0.47}          & 0.58          \\ \cline{2-10} 
                                                                                & 7                             & \multicolumn{1}{c|}{0.1}                & \multicolumn{1}{c|}{0.1}                   & \multicolumn{1}{c|}{0.2}           & 0.39          & \multicolumn{1}{c|}{0.23}               & \multicolumn{1}{c|}{0.1}                   & \multicolumn{1}{c|}{0.47}          & 0.66          \\ \cline{2-10} 
                                                                                & 9                             & \multicolumn{1}{c|}{0.24}               & \multicolumn{1}{c|}{0.1}                   & \multicolumn{1}{c|}{1.57}          & 1.53          & \multicolumn{1}{c|}{0.29}               & \multicolumn{1}{c|}{0.1}                   & \multicolumn{1}{c|}{0.47}          & 0.69          \\ \hline
\end{tabular}
\end{table*}

\paragraph*{Limitation and future works} The sensor is tested for low-frequency contacts and cannot handle high-frequency contacts. We will look into novel structure designs and materials to cope with this limitation. While our current calibration is focused on measuring normal forces, we are exploring the skin's capabilities to measure shear force and torque along with multi-point contact localisation. Moreover, from an application point of view, we plan to develop a miniature version of the skin to enable real-time force control for a strawberry harvesting robotic end effector~\cite{10053882}. Future works include testing the sensor in soft and deformable object manipulation~\cite{nazari2023deep}. 
\section{Conclusion}
This paper introduces a new design and generalisable calibration approach of the AST skin, an innovative tactile technology. AST skin utilises acoustic channels beneath its surface to modulate acoustic waves in response to external stimuli. Machine learning techniques interpret these modulations as tactile measurements, offering adaptability for diverse applications. The standalone module of AST technology separates the sensing skin from the acoustic hardware, enabling easy reshaping to meet specific needs. Notably, our previous examinations confirmed AST Skin's resilience to external sound disturbances, underscoring its robustness and reliability~\cite{vishnu2023acoustic}. Our study focused on assessing the impact of single and double layers of acoustic channels on sensing performance. We found that the bi-layered design consistently showed an all-around performance in measuring geometric features, contact locations, and normal force. Moreover, our study illustrated the skin's capacity to derive tactile features even from non-calibrated sensing points, thereby portraying it as a continuous sensing area.


\bibliographystyle{IEEEtran}

\bibliography{IEEEexample,Reference}

\end{document}